\documentclass[10pt,twocolumn,letterpaper]{article}

\usepackage[pagenumbers]{cvpr} 

\definecolor{cvprblue}{rgb}{0.21,0.49,0.74}
\usepackage[pagebackref,breaklinks,colorlinks,allcolors=cvprblue]{hyperref}
\usepackage{bm}
\usepackage{subcaption}

\usepackage{amssymb}
\usepackage{pifont}
\newcommand{\cmark}{\color{teal}\ding{51}}%
\newcommand{\xmark}{\color{red}\ding{55}}%

\title{IQViC: In-context, Question Adaptive Visual Compressor\\for Long-term Video Understanding LMMs}

\author{
    Sosuke Yamao\thanks{These authors contributed equally to this work} \quad 
    Natsuki Miyahara\footnotemark[1] \quad 
    Yuki Harazano \quad 
    Shun Takeuchi\\
    Fujitsu Research, Fujitsu Limited \\
{\tt\small \{yamao.sosuke, n-miyahara, harazono.yuki, takeuchi.shun\}@fujitsu.com}
}

\begin{document}
\maketitle
\begin{abstract}

With the increasing complexity of video data and the need for more efficient long-term temporal understanding, 
existing long-term video understanding methods often fail to accurately capture and analyze extended video sequences. 
These methods typically struggle to maintain performance over longer durations 
and to handle the intricate dependencies within the video content.
To address these limitations, 
we propose a simple yet effective large multi-modal model framework for long-term video understanding
that incorporates a novel visual compressor, the In-context, Question Adaptive Visual Compressor (IQViC).
The key idea, inspired by humans' selective attention and in-context memory mechanisms,
is to introduce a novel visual compressor and incorporate efficient memory management techniques
to enhance long-term video question answering.
Our framework utilizes IQViC, a transformer-based visual compressor, 
enabling question-conditioned in-context compression, 
unlike existing methods that rely on full video visual features. 
This selectively extracts relevant information, 
significantly reducing memory token requirements.
Through extensive experiments on a new dataset based on InfiniBench for long-term video understanding,
and standard benchmarks used for existing methods' evaluation,
we demonstrate the effectiveness of our proposed IQViC framework and its superiority over state-of-the-art methods 
in terms of video understanding accuracy and memory efficiency.

\end{abstract}

\section{Introduction}
\label{sec:intro}

\begin{figure}[!t] 
  \begin{minipage}[b]{1.0\hsize}
    \centering
    \includegraphics[width=1.0\linewidth]{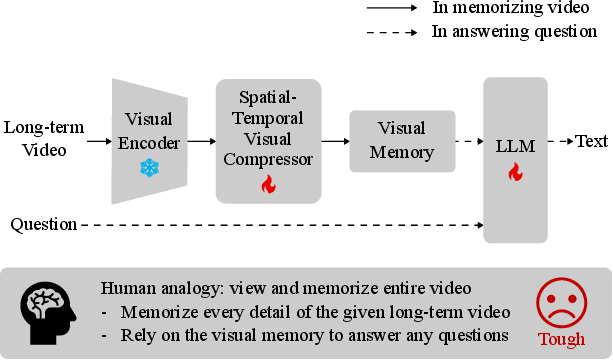}
    \subcaption{Conventional methods}
    \vspace{0.4cm}
    \label{fig:overview-a}
  \end{minipage}
  \begin{minipage}[b]{1.0\hsize}
    \centering
    \includegraphics[width=1.0\linewidth]{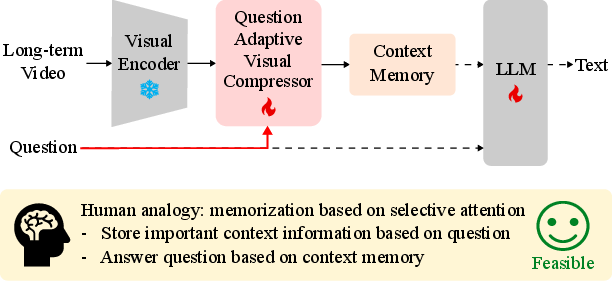}
    \subcaption{Proposed method}
    \label{fig:overview-b}
  \end{minipage}
  \caption{Comparison of conventional and proposed methods for long-term video understanding.
  (a)~Conventional methods store temporally and spatially compressed image features in the memory. 
  In human analogy, the architecture remembers everything in a video and then answers any question, which is a tough task.
  (b)~The proposed method compresses the visual information into context conditioned by the text prompt. 
  In human analogy, this architecture focuses on the necessary information with selective attention while watching a video based on a question 
  and then answers the question, which is a feasible task.
  }
  \label{fig:overview} 
\end{figure}

Recent breakthroughs in large language models (LLMs)~\cite{NEURIPS2020_1457c0d6, Achiam2023GPT4TR, Touvron2023LLaMAOA, Touvron2023Llama2O, Reid2024Gemini1U} 
have demonstrated impressive capabilities, suggesting various 
potential applications~\cite{Minaee2024LargeLM, Zhao2023ASO, HadiLargeLM, Gao2023RetrievalAugmentedGF, Li2023LargeLM, Kasneci2023ChatGPTFG}.
The development of visual encoders~\cite{Dosovitskiy2020AnII} and modality alignment techniques~\cite{Li2023BLIP2BL, Dai2023InstructBLIPTG} has significantly improved the ability of LLMs to perform cross-modal tasks. This is particularly evident in the field of visual question answering (VQA)~\cite{khan2022transformers, tsimpoukelli2021multimodal}, where LLMs now exhibit exceptional performance in understanding images and videos.
As evidenced by recent technological trends~\cite{tang2024videounderstandinglargelanguage, zou2024secondshoursreviewingmultimodal, yin2023survey}, ongoing research in VQA has proposed improved methodologies and benchmarks, leading to demonstrable enhancements 
in the performance of image and video understanding using LLMs~\cite{zhang2024vision, tang2024videounderstandinglargelanguage}.

Although high performance has been demonstrated in understanding images and short videos, the long video understanding task remains challenging.
The primary challenge is to process the vast amount of tokens generated from a long-term video without exceeding the inferent context window limitation of LLMs and GPU memory capacity.
To address this challenge, some recent studies have proposed memory-augmented LLM frameworks, such as MovieChat~\cite{Song2023MovieChatFD}, MA-LMM~\cite{He2024MALMMML}, and Flash-VStream~\cite{Zhang2024FlashVStreamMR},
which store visual features of video frames in a short-term or long-term memory bank and generate answers based on the stored features and the input question.
These strategies, however, often lead to redundant information storage because they try to retain all the visual information of the video frames, which is inefficient with limited memory resources.
Moreover, since these methods compress visual features spatially or temporally with simple weighted averaging or pooling operations,
they suffer from significant information loss and smoothing artifacts, thereby reducing the accuracy of LLM responses.

To address the limitations of existing methods, we draw inspiration from human memory and selective attention mechanisms.
Human memory is not a simple spatio-temporal record~\cite{Cowan1988EvolvingCO}; rather, it is selective, prioritizing information relevant to a given theme or task, as illustrated by the ``Invisible Gorilla" experiment~\cite{Simons1999GorillasIO}. 
This selective attention, where relevant information is retained and irrelevant information is suppressed, suggests an analogous and promising approach for efficient long-term video understanding. 
Therefore, effective processing requires memory mechanisms that incorporate video context and selectively retain crucial information as context in light of the question.

With these in mind, we propose an {\bf I}n-context, {\bf Q}uestion adaptive {\bf Vi}sual {\bf C}ompressor (IQViC) for long-term video understanding large multi-modal models (LMMs). 
IQViC shares a similar architecture with the existing memory-augmented LLMs designed for long-term video understanding~\cite{He2024MALMMML, Song2023MovieChatFD, Zhang2024FlashVStreamMR}. 
As shown in Figure~\ref{fig:overview}, unlike conventional approaches that use spatial-temporal compressor to memorize all the visual information in the video,
IQViC incorporates a question adaptive visual compressor to store the essential information based on the input question.
The proposed visual compressor selectively attends to information relevant to the posed question with transformer-based encoding,
resulting in substantial compression by reducing the number of tokens. 
By discarding redundant information and retaining only the essential features, 
the proposed approach effectively reduces the memory consumption without compromising the accuracy of the LLM's responses.

We summarize our main contributions as follows:

\begin{itemize}
  \item 
  We propose a visual compressor that can extract important information (i.e., context)
  from a video frame in a lightweight manner, adapting to the given question, 
  inspired by the humans' selective attention and in-context memory mechanisms.
  \item 
  We propose a new LMM framework for long-term video understanding 
  that incorporates the proposed visual compressor, which spatially compresses visual information as context based on the question,
  and the context memory mechanism, which temporally compresses the extracted context to maintain memory efficiency.
  \item 
  We conduct extensive quantitative evaluations from the perspectives of both long-term and short-term video understanding.
  Our results demonstrate that the proposed IQViC framework outperforms state-of-the-art methods in terms of both performance and memory efficiency. 
\end{itemize}

\section{Related Work}
\label{sec:relatedwork}

\subsection{Large Multi-modal Models}
\label{subsec:large-multi-modal-models}
The remarkable progress of LLMs and natural language processing (NLP) has been driven by the development of numerous architectural frameworks, 
leading to the emergence of cross-modal models~\cite{Zhou2023ACS, Huang2022TowardsRI}.
For instance, the BLIP series~\cite{Li2022BLIPBL, Li2023BLIP2BL, Dai2023InstructBLIPTG} incorporates a Querying-transformer to bridge the gap between frozen image encoders and frozen LLMs, 
resulting in a model with significantly fewer trainable parameters than previous approaches.
LLaVA series~\cite{liu2023improvedllava,Liu2023VisualIT} was developed by integrating 
CLIP~\cite{Radford2021LearningTV} pre-trained vision encoder and Vicuna~\cite{vicuna2023} 
language decoder, trained on a newly constructed dataset of language-image pairs 
designed for instruction-following tasks.
The models proposed in~\cite{liu2024llavanext, Jin2023ChatUniViUV, Cheng2024VideoLLaMA2A, Li2023LLaMAVIDAI, Maaz2023VideoChatGPTTD}, 
including those that incorporate the visual-language connector described earlier, are capable of handling multiple images and videos. 
However, our approach introduces an LLM framework designed for significantly longer video sequences, 
exceeding the temporal limitations of these prior methods.

\begin{figure*}[!t]
  \centering
  \includegraphics[width=1.0\linewidth]{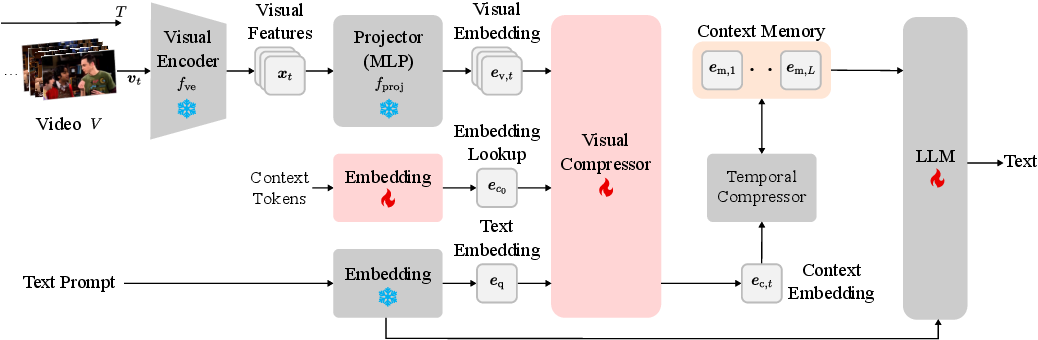}
  \caption{The architecture of the proposed IQViC framework in detail. 
  The visual encoder extracts visual features from each video frame and projects them into visual embeddings.
  Then, the proposed visual compressor, IQViC, transforms the visual embeddings into context embeddings conditioned on the text prompt.
  The context memory stores the compressed context embeddings sequentially with a predefined capacity.
  Finally, the LLM generates answer to the text prompt.
  The snowflake symbol represents components with fixed parameters, while the flame icon indicates the modules that are trained.
  }
  \label{fig:overview_detail}
\end{figure*}

\subsection{Long-term Video Understanding}
\label{subsec:long-term-video-understanding}
Understanding long-term videos presents significant challenges due to high computational costs and the risk of overlooking crucial information inherent in their extended duration. 
Conventional video understanding methods~\cite{Maaz2023VideoChatGPTTD, Li2023VideoChatCV, Zhang2023VideoLLaMAAI, Li2023LLaMAVIDAI}, 
are typically designed for short clips, which are generally less than 1 minute in length. These methods are not suitable for long videos.
To address this limitation, various frameworks for efficient processing of long videos have been proposed. 
Herein, we focus on ``very-long" videos, defined as videos exceeding 10 minutes in length~\cite{zou2024secondshoursreviewingmultimodal}.

Recently, several studies have proposed approaches to address the challenges of long-term video understanding~\cite{zou2024secondshoursreviewingmultimodal, Zhang2024LongCT, Lin2023MMVIDAV, Qiu2024ArtemisTR}.
LongVA~\cite{Zhang2024LongCT} achieves efficient long-term video processing through long context transfer, 
which involves context length extention of the underlying LLM and 
modality alignment to adapt the extended context to long video sequences.
MM-Vid~\cite{Lin2023MMVIDAV} performs advanced video understanding through 
the synergistic combination of GPT-4V(ision)'s capabilities and specialized tools 
for video, audio, and automatic speech recognition.
Artemis~\cite{Qiu2024ArtemisTR} addresses the challenge of referential understanding 
within complex video data by leveraging video, a text prompt, and a bounding box 
annotation specifying a region of interest within an arbitrarily selected frame, and performs 
comprehensive description of the referent's presence and actions throughout the duration of the video.

Moreover, some studies have introduced memory-augmented frameworks for video understanding, 
MA-LMM~\cite{He2024MALMMML} addresses the challenge by processing videos online and storing past information in a long-term memory bank, incorporating a compression method allowing it to reference historical content without exceeding LLM constraints.
MovieChat~\cite{Song2023MovieChatFD} leverages a memory model, using transformer tokens as memory carriers and a specially designed memory mechanism with short-term and long-term components. 
Flash-VStream~\cite{Zhang2024FlashVStreamMR} leverages a multi-memory named Spatial-Temporal-Abstract-Retrieved (STAR) memory to efficiently compress visual information, significantly reducing inference latency and VRAM consumption compared to existing approaches. 

Although memory augmentation is a well-established approach, a significant limitation arises from the smoothing effect of content-agnostic compression methods, which can severely degrade memory fidelity. 
Therefore, this study proposes a novel memory mechanism that incorporates in-context compression.

\subsection{In-context Compression}
\label{subsec:context-compression}
Recent research has highlighted the potential of LLMs as powerful general-purpose compression algorithms, suggesting a new paradigm for compression technology~\cite{Deletang2023LanguageMI}.
In particular, research on prompt compression for Question Answering (QA) focuses on compressing long contexts into shorter representations using pre-trained LLMs.
GIST~\cite{Mu2023LearningTC}, although a prompt compression method, suffers from the limitation that its compression target does not encompass the long contexts typically considered in our work.
In contrast, Autocompressor~\cite{Chevalier2023AdaptingLM} compresses long contexts by segmenting the entire context and recursively generating summary vectors.
In-context Autoencoder (ICAE)~\cite{Ge2023IncontextAF} shares conceptual similarities with GIST and Autocompressor, but it offers advantages, in terms of both compression efficiency and architectural simplicity. 
It achieves higher compression ratios for long contexts and demonstrates versatility across diverse tasks due to its pre-trained autoencoder. 
However, its current limitation is its exclusive use of text modality, presenting a challenge in terms of its application to visual data.
Therefore, we propose an ICAE-inspired VQA framework incorporating a visual modality encoder conditioned on text prompts, which we call visual compressor.

\section{Method}

We introduce IQViC, an in-context, question adaptive visual compressor for long-term video understanding.
Unlike conventional methods that store temporally or spatially compressed image features 
in memory~\cite{Song2023MovieChatFD, He2024MALMMML, Zhang2024FlashVStreamMR}, 
we propose a framework incorporating a visual compressor that compresses image features to compact context 
conditioned by the text prompt and a context memory that stores the compressed context in a sequential manner.
The design of the framework draws inspiration from memory and selective attention mechanisms in humans,
which prioritize and retain relevant information to a given task while suppressing irrelevant information~\cite{Cowan1988EvolvingCO, Simons1999GorillasIO}.
Figure~\ref{fig:overview_detail} presents an overview of the IQViC framework, 
which comprises four main components: (i)~visual encoder, (ii)~visual compressor, (iii)~context memory, and (iv)~decoder.
The following sections describe each part in detail, the training method, and the inference method of the proposed framework.

\subsection{Visual Encoder}
Given a sequence of $T$~video frames $V = [\bm{v}_1, \bm{v}_2, ..., \bm{v}_T]$, 
each frame $\bm{v}_t \in \mathbb{R}^{H \times W \times 3}$ on the frame index $t$
is encoded to visual feature 
using a pre-trained visual encoder $f_{\rm{ve}}$ as follows:
\begin{equation}
  \label{visencoder}
  \bm{x}_t = f_{\rm{ve}}(\bm{v}_t) \in \mathbb{R}^{P \times D_{\rm{f}}},
\end{equation}
where $H$ and $W$ are the height and width of the frame, respectively, 
and $P$ is the number of patch tokens, with $D_{\rm{f}}$ being the dimension of each token.
We use CLIP~ViT-L/14-336px~\cite{Radford2021LearningTV} as the visual encoder.
The visual feature $\bm{x}_t$ is passed to a pre-trained visual projector for modality alignment, 
generating visual embeddings for the next step:
\begin{equation}
  \label{projector}
  \bm{e}_{{\rm{v}},t} = f_{\rm{proj}}(\bm{x}_t) \in \mathbb{R}^{P \times D_{\rm{e}}},
\end{equation}
where $D_{\rm{e}}$ is the dimension of the embedding.
We use the two-layer multi-layer perceptron (MLP) from
LLaVA-v1.5-7B~\cite{liu2023improvedllava} for the visual projector.

\subsection{Visual Compressor}
We propose a visual compressor that outputs a compact context embedding conditioned by image features 
and a text prompt to compress the information necessary for answering a question to achieve accurate video understanding with compact memory.
Inspired by ICAE~\cite{Ge2023IncontextAF}, we use a transformer-based LLM for the visual compressor
to compress image features into lightweight context embeddings associated with the text prompt 
through the self-attention mechanism of the transformer architecture.
By referencing the transformer's hidden states of the last layer, we extract a compressed contextual representation as the result of the compression.

The input of the visual compressor is a concatenation of the visual embedding $\bm{e}_{{\rm{v}},t}$ 
and text prompt embedding $\bm{e}_{\rm{q}} \in \mathbb{R}^{K \times D_{\rm{e}}}$, 
augmented by a learnable embedding lookup of the context token 
$\bm{e}_{\rm{c_0}} \in \mathbb{R}^{C \times D_{\rm{e}}}$
to obtain their outputs as the context to memorize for the visual and the text prompt:
\begin{equation}
  \label{prepareinput}
  \bm{e}_{\rm{enc,in}} = \mathtt{Concat}[\bm{e}_{\rm{q}},\ \bm{e}_{{\rm{v}},t},\ \bm{e}_{\rm{c_0}}],
\end{equation}
where $K$ and $C$ $(\ll P)$ are the number of tokens.
We extract the context embedding $\bm{e}_{{\rm{c}},t} = \bm{e}'_{\rm{c_0}} \in \mathbb{R}^{C \times D_{\rm{e}}}$ from the output of the visual compressor $f_{\rm{enc}}$ as: 
\begin{equation}
  \label{viscompressor}
  \bm{e}_{\rm{enc,out}} = f_{\rm{enc}}(\bm{e}_{\rm{enc,in}}) = \mathtt{Concat}[\bm{e}'_{\rm{q}},\ \bm{e}'_{{\rm{v}},t},\ \bm{e}'_{\rm{c_0}}],
\end{equation}
where $\bm{e}'_{\rm{q}} \in \mathbb{R}^{K \times D_{\rm{e}}}$ 
and $\bm{e}'_{{\rm{v}},t} \in \mathbb{R}^{P \times D_{\rm{e}}}$
are the corresponding output embeddings of the text prompt and the visual feature, respectively.

\begin{figure*}[!t] 
  \begin{minipage}[b]{0.46\hsize}
    \centering
    \includegraphics[width=1.0\linewidth]{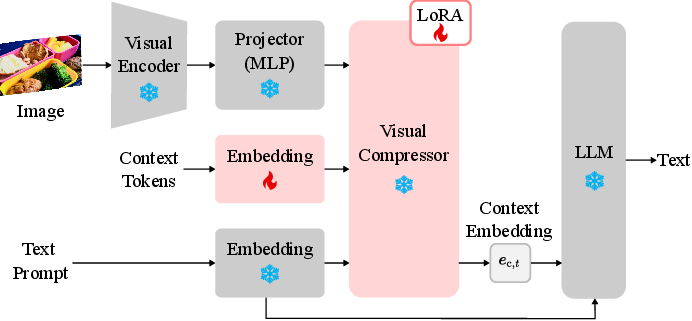}
    \subcaption{Step~1: Training the visual compressor and embedding module}
    \label{fig:training_method-a}
  \end{minipage} 
  \begin{minipage}[b]{0.54\hsize}
    \centering
    \includegraphics[width=1.0\linewidth]{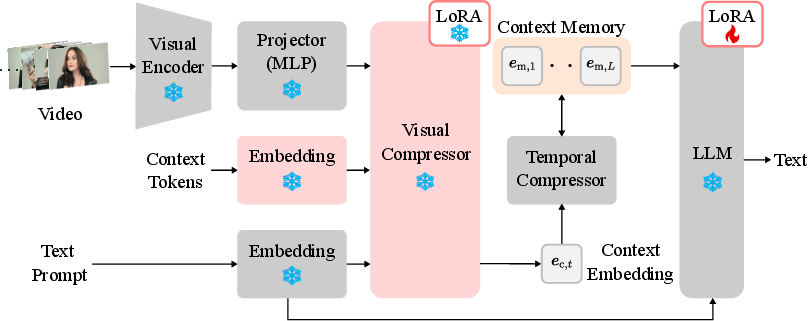}
    \subcaption{Step~2: Finetuning the LLM}
    \label{fig:training_method-b} 
  \end{minipage}
  \caption{Training method of the proposed IQViC framework in two steps. 
  The snowflake symbol represents modules with fixed parameters, while the flame icon represents the trainable parts.
  (a)~In Step~1, the visual compressor with LoRA and embedding module are trained using image QA datasets.
  (b)~In Step~2, we finetune the LoRA-adapted LLM using the modules trained in Step 1, using a video QA dataset.}
  \label{fig:training_method} 
\end{figure*}

\subsection{Context Memory}
Context memory $\bm{e}_{\rm{m}} = [\bm{e}_{{\rm{m}},1},\ \bm{e}_{{\rm{m}},2},\ ...,\ \bm{e}_{{\rm{m}},L}]$ 
stores context embeddings obtained from the visual compressor in a sequential manner 
with a predefined capacity $L$.
If the addition of a new context embedding causes the memory to exceed its capacity, 
a temporal compressor is applied to compress the stored embeddings to fit within the capacity, 
resulting in an updated context memory.
Similar to the work in~\cite{Song2023MovieChatFD, He2024MALMMML}, 
the temporal compressor aggregates and compresses context embeddings over time 
by leveraging the similarity between adjacent features. 
When a new context embedding $e_{{\rm{m}},L+1}$ comes in, 
the temporal compressor calculates the cosine similarity between all the adjacent embeddings, as: 
\begin{equation}
  \label{memory_calc_similarity}
  s_i = \cos(\bm{e}_{{\rm{m}},i},\ \bm{e}_{{\rm{m}},i+1}),\ i \in [1, L].
\end{equation}
Then selects the pair with the highest similarity and averages the selected embeddings:
\begin{align}
  \label{memory_update}
  k &= \underset{i} {\operatorname{argmax}}\ s_i, \\
  \hat{\bm{e}}_{{\rm{m}},k} &= (\bm{e}_{{\rm{m}},k} + \bm{e}_{{\rm{m}},k+1})/2.
\end{align}
This parameter-free algorithm readily integrates into frame-based visual encoders. 
The added computational cost of frame similarity is negligible compared to the efficiency improvements from reduced frame storage.

\subsection{Decoding}
Finally, the decoder generates a response to the text prompt. 
Same as the visual compressor, we use a transformer-based LLM for the decoder.
This is achieved by concatenating the text prompt embedding with the context memory constructed in the previous step as:
\begin{equation}
  \label{decoderinput}
  \bm{e}_{\rm{dec,in}} = \mathtt{Concat}[\bm{e}_{\rm{q}},\ \bm{e}_{\rm{m}}],
\end{equation}
and feeding it to the decoder $f_{\rm{LLM}}$ and obtaining text output $a$ as:
\begin{equation}
  \label{decoding}
  a = f_{\rm{LLM}}(\bm{e}_{\rm{dec,in}}).
\end{equation}

\subsection{Training Method}

We train the visual compressor, context embedding lookup, and LLM of the IQViC framework. 
Ideally, all modules would be trained jointly end-to-end on image and video QA datasets for optimal performance.
However, since the architecture accumulates the output of the visual compressor into the context memory sequentially,
we found that training with video frames resulted in an unmanageable memory consumption for backpropagation even if we used LoRA~\cite{Hu2021LoRALA} to reduce the memory footprint.
Consequently, we employ a two-step training method for the IQViC framework, as shown in Figure~\ref{fig:training_method}.

In Step 1, we train the visual compressor and embedding module using image QA datasets, as illustrated in Figure~\ref{fig:training_method-a}. 
The training data comprise the standard image QA datasets including 
COCO~\cite{Lin2014MicrosoftCC}, GQA~\cite{Hudson2019GQAAN}, TextVQA~\cite{Singh2019TowardsVM}, OCR-VQA~\cite{Mishra2019OCRVQAVQ}, 
and Visual~Genome~\cite{Krishna2016VisualGC}, defined as the annotation data LLaVA-filtered-665K~\cite{Liu2023VisualIT},
with the finetuned Vicuna-v1.5~\cite{vicuna2023} based on LLaVA-v1.5-7B~\cite{liu2023improvedllava} serving as the visual compressor.
This step focuses on learning to compress video frames into a lightweight contextual representation.

Step 2 involves training the LLM with the frozen visual compressor and embedding model from Step 1, as shown in Figure~\ref{fig:training_method-b}.
The training data consists of video QA pairs from ActivityNet~\cite{Heilbron2015ActivityNetAL}, 
defined as Video-ChatGPT-filtered-98K~\cite{Zhang2024FlashVStreamMR}, 
again using the finetuned Vicuna-v1.5 based on LLaVA-v1.5-7B as the base model. 
This step aims to learn QA capabilities that reference the context memory.

Table~\ref{tab:hyperparameter} presents the hyperparameters used in Steps 1 and 2 of IQViC framework training.
Similar to ICAE~\cite{Ge2023IncontextAF}, we finetune LoRA-adapted LLMs for the visual compressor and the decoder.
LoRA is applied to the query and value projections of the multi-head self-attention layers of the transformer.  
During training, the context token $C=64$ and the capacity of the context memory $L=10$ are set to balance performance and resource consumption.
Our model is supervised with the standard cross-entropy loss for both steps.
Every training is conducted on eight~A100 40GB GPUs. 
Step 1 takes approximately 12~hours, whereas Step 2 takes approximately 48~hours.

\begin{table}[!t]
  \centering
  \caption{Hyperparameters used in the training of the IQViC framework.
  We use the same hyperparameters for both training steps.}
  \begin{tabular}{@{}lcc@{}}
      \toprule
      {\bf Hyperparameter} & {\bf Value}\\
      \midrule
      Optimizer & AdamW~\cite{Loshchilov2017DecoupledWD} \\
      Batch size & 4 \\
      Gradient accumulation steps & 4 \\
      Learning rate & $2 \times 10^{-4}$ \\
      LoRA rank & 64 \\
      LoRA alpha & 16 \\
      LoRA dropout & 0.05 \\
      Context token $C$ & 64 \\
      Context memory capacity $L$ & 10 \\
      \bottomrule
  \end{tabular}
  \label{tab:hyperparameter}
  \end{table}

\subsection{Inference}

During inference, 
video frames are first sequentially compressed using a visual compressor, 
conditioned on the provided text prompt, to construct the context memory. 
This context memory, along with the text prompt, 
is then provided as input to the decoder to generate the text response. 
This approach enables typical batch inference on the entire video and, 
similar to Flash-VStream~\cite{Zhang2024FlashVStreamMR}, 
also enables online QA through the parallelization of 
context memory construction and response generation.

\section{Experiments}
We conduct quantitative evaluations of the proposed IQViC framework in comparison to 
the state-of-the-art methods on long-term and short-term VQA benchmarks.
Additionally, we perform an ablation study to analyze the effectiveness of the proposed method.

\begin{table*}[!t]
    \centering
    \caption{Comparison of memory architectures and quantitative evaluation for long-term video QA on InfiniBench-Vision.
    The ``Memory Modules" column indicates the memory modules included in each method.
    The ``Memory Tokens" column indicates the total number of tokens in a full memory after a long-term video input; 
    parenthetical values show the token count per module.
    Acc. and Sco. denote accuracy and score, respectively.
    The best results (i.e., lowest token count, highest accuracy, and highest score) are highlighted in bold.}
    \begin{tabular}{@{}lcccc@{}}
        \toprule
        {\bf Method} & {\bf Memory Modules} & {\bf Memory Tokens} & {\bf Acc.} & {\bf Sco.} \\
        \midrule
        MovieChat~\cite{Song2023MovieChatFD} & Short-term / Long-term memory & 8768 $(576+8192)$  & 14.2 & 1.2 \\ %
        MA-LMM~\cite{He2024MALMMML} & Visual / Query (12-layer) memory & 12800 $(5120+7680)$ & 42.6 & 2.4 \\
        Flash-VStream~\cite{Zhang2024FlashVStreamMR} & Spatial / Temporal / Abstract / Retrieved memory & 681 $(64+400+25+192)$ & 44.3 & 2.4 \\
        \midrule
        {\bf IQViC} & Context memory & {\bf 640} & {\bf 49.1} & {\bf 2.6} \\
        \bottomrule
    \end{tabular}
    \label{tab:Long-term VideoQA}
    \end{table*}

\subsection{Setup}

\paragraph{Datasets.}

To evaluate the long-term video understanding capability of the IQViC framework,
we consider adopting InfiniBench~\cite{Ataallah2024InfiniBenchAC}, the challenging long-term video QA dataset.
InfiniBench provides a large-scale video QA dataset of drama TV series and movies, 
with an average length of 52.6 minutes and 108.2K QA pairs. 
However, we found that most QA pairs are difficult answer with video frames only, 
which require subtitles, audio, or prior knowledge of the contents.
To align with the conditions of existing zero-shot long-term video QA benchmark datasets\!\!~\footnote{Due to licensing restrictions, InfiniBench is the only long-term video understanding benchmark dataset accessible to us.}
like MLVU~\cite{Zhou2024MLVUAC} and LVBench~\cite{Wang2024LVBenchAE},
we carefully filter the questions and create a new subset of InfiniBench called {\bf InfiniBench-Vision}.
In establishing InfiniBench-Vision, following the annotation policy in MLVU, 
we assume that questions containing person names that are specific to the content are difficult to answer with video frames only
and we remove the hard questions from the original InfiniBench using GPT-4.
InfiniBench-Vision contains 265 videos with an average length of 49.0 minutes and 599 open-ended QA pairs.
For details of the filtering process and dataset statistics, refer to the supplementary material.

To evaluate the basic short-term video QA capability of the IQViC framework, 
we conduct zero-shot open-ended video QA experiments on standard benchmarks,
NExT-QA~\cite{Xiao2021NExTQANP}, MSVD-QA~\cite{Xu2017VideoQA}, and MSRVTT-QA~\cite{Xu2017VideoQA}.
These benchmarks consist of short-term videos with durations ranging from approximately 10 to 40 seconds.
In the ablation study, we use challenging image QA dataset LLaVA-Bench (in-the-wild)~\cite{Liu2023VisualIT}
to evaluate the effectiveness of the proposed visual compressor.

\paragraph{Evaluation metrics.}

We use GPT to calculate the accuracy and score following the common practices 
in~\cite{Song2023MovieChatFD, He2024MALMMML, Zhang2024FlashVStreamMR, Ataallah2024InfiniBenchAC}.
For each question, we provide the prediction and the ground truth answer to GPT,
and ask it to determine whether the prediction is correct and to provide a score between 0 and 5.
We then calculate the accuracy and the average score for all questions.
The accuracy is calculated as the ratio of the number of correct predictions to the number of questions.
We use GPT-4 on InfiniBench-Vision and LLaVA-Bench (in-the-wild), 
and GPT-3.5 on NExT-QA, MSVD-QA, and MSRVTT-QA to ensure fair comparison with the existing methods.

\subsection{Zero-shot Video Question Answering}

\begin{table}[!t]
    \small
    \centering
    \setlength{\tabcolsep}{5.3pt}
    \caption{Quantitative evaluation for short-term video QA.
    Acc. and Sco. denote accuracy and score, respectively.
    The best score is highlighted in bold, and the second-best score is underlined. 
    *:~Evaluated by us with the official implementation and pre-trained model provided by the authors.}
    \begin{tabular}{@{}lcccccc@{}}
        \toprule
        {\bf Method} &
        \multicolumn{2}{c}{{\bf NExT-QA}} & 
        \multicolumn{2}{c}{{\bf MSVD-QA}} & 
        \multicolumn{2}{c}{{\bf MSRVTT-QA}} \\
        ~ & {\bf Acc.} & {\bf Sco.} & {\bf Acc.} & {\bf Sco.} & {\bf Acc.} & {\bf Sco.} \\
        \midrule
        MovieChat & {\bf 49.9} & 2.7 & {\bf 75.2} & 3.8 & 52.7 & 2.6 \\ 
        MA-LMM & \text{-} & \text{-} & 60.6 & \text{-} & 48.5 & \text{-} \\
        Flash-VStream* & 47.9 & \underline{3.3} & 69.6 & \underline{3.9} & \underline{55.2} & \underline{3.4} \\
        \midrule
        {\bf IQViC} & \underline{49.8} & {\bf 4.0} & \underline{72.4} & {\bf 4.0} & {\bf 59.5} & {\bf 3.5} \\
        \bottomrule
    \end{tabular}
    \label{tab:Short-term VideoQA}
    \end{table}

\paragraph{Long-term video question answering.}

A comparative analysis of the proposed IQViC framework against state-of-the-art methods—MovieChat~\cite{Song2023MovieChatFD}, MA-LMM~\cite{He2024MALMMML}, 
and Flash-VStream~\cite{Zhang2024FlashVStreamMR}—is presented in Table \ref{tab:Long-term VideoQA}. 
This analysis is conducted on the InfiniBench-Vision dataset using the official implementations 
and pre-trained models provided by the respective authors.
Our proposed IQViC achieves the highest average score, surpassing existing state-of-the-art methods in long-term video understanding.
Specifically, it demonstrates a 4.8\% accuracy improvement over Flash-VStream, the previous top performer. 
More importantly, this superior performance is achieved with the simplest memory architecture 
and the most efficient memory token utilization among the evaluated methods.

The proposed visual compressor, leveraging the framework's concept of selective information 
compression, was expected to improve accuracy with reduced memory consumption compared to 
conventional methods that process all visual information. 
The results strongly support the hypothesis: IQViC demonstrated superior VQA performance 
while minimizing memory usage, exceeding the capabilities of previous frameworks.

\begin{table}[!t]
    \centering
    \caption{Ablation study using visual compressor on image QA dataset LLaVA-Bench (in-the-wild)~\cite{Liu2023VisualIT}.
    ``Vis. Input" denotes the token size of the input visual information to the decoder.
    The compression ratio {\bf \(r_C\)} is calculated as \(r_C = C/P\) in percentage,
    where \(C\) is the number of context tokens and \(P\) is the number of patch tokens of the visual embeddings (i.e., 576 tokens).}
    \begin{tabular}{@{}lcccc@{}} 
        \toprule
        {\bf Method} & {\bf Vis. Input} & {\bf \(r_C\)} & {\bf Acc.} \\
        \midrule
        Flash-VStream & 145 & 25\% & 47.8 \\ 
        \midrule 
        Avg Pool (\(C=64\))  & 64 & 11\% & 45.0 \\
        IQViC (\(C=1\))   & 1 & 0.2\% & 49.9 \\ 
        IQViC (\(C=32\))  & 32 & 5.6\% & 55.2 \\ 
        IQViC (\(C=64\))  & 64 & 11\% & {\bf 55.4} \\
        \bottomrule
    \end{tabular}
    \label{tab:Ablation study image answering}   
    \end{table}

\paragraph{Short-term video question answering.}
Table \ref{tab:Short-term VideoQA} compares the proposed IQViC framework with state-of-the-art methods on three short-term VQA benchmarks,
NExT-QA, MSVD-QA, and MSRVTT-QA, to evaluate its performance in open-ended video QA tasks.
Notably, the results demonstrate that IQViC achieves state-of-the-art performance according to 
the scores obtained on all datasets, and the proposed method improves upon the previous best method 
in MSRVTT-QA, Flash-VStream by 4.3\% in accuracy.
Although the proposed framework, designed for long-term video understanding, 
is expected to show reduced effectiveness on short-term video datasets, it achieved the best performance, nonetheless.
Despite being designed specifically for long-term videos, the proposed demonstrates 
the robustness to short-term videos, with experimental results showing no performance degradation. 

\subsection{Ablation Study}

\paragraph{Compressor comparison.}
We conduct a detailed evaluation of the visual compressor's compression performance using a simplified, 
single-image input architecture (Figure~\ref{fig:training_method-a}).
The evaluation employs the LLaVA-Bench~(in-the-wild), a challenging image understanding benchmark dataset. 
Although it contains only images, this dataset offers substantial content variability, presenting a rigorous test for image understanding models.
Table~\ref{tab:Ablation study image answering} presents a performance comparison 
against other compression techniques to evaluate the efficacy of visual compressor. 
Maintaining the proposed architecture, we substitute the visual compressor with average pooling 
and compare the performance of our in-context compression method with standard feature compression 
at an equivalent compression ratio.
The results demonstrate that our visual compressor achieves a performance improvement of more than 10\% over average pooling with compression ratios of 11\% and 5.6\%. 
This improvement is attributed to IQViC's ability to selectively compress data, removing redundancy 
while retaining necessary information, leading to substantially greater efficiency compared to 
standard compression techniques.

\paragraph{Context token size ablation.}
Table \ref{tab:Ablation study image answering} also shows the results of a hyperparameter study performed using the LLaVA-Bench~(in-the-wild) dataset. 
The study investigates the influence of the number of context tokens on VQA performance.
As expected, reducing the number of context tokens results in decreased performance due to limitations in expressiveness. 
However, the proposed method maintains a surprisingly high accuracy of 49\% even with only one context token, 
thereby outperforming Flash-VStream by 145 tokens. 
This unexpected result, consistent with our observations in short-term video understanding evaluations, 
underscores the effectiveness of the visual compressor in image understanding. 
Future research will focus on further investigating the representation capabilities of context tokens.

\subsection{Case Study}
\begin{figure*}[t]
    \centering
    \includegraphics[width=1.0\linewidth]{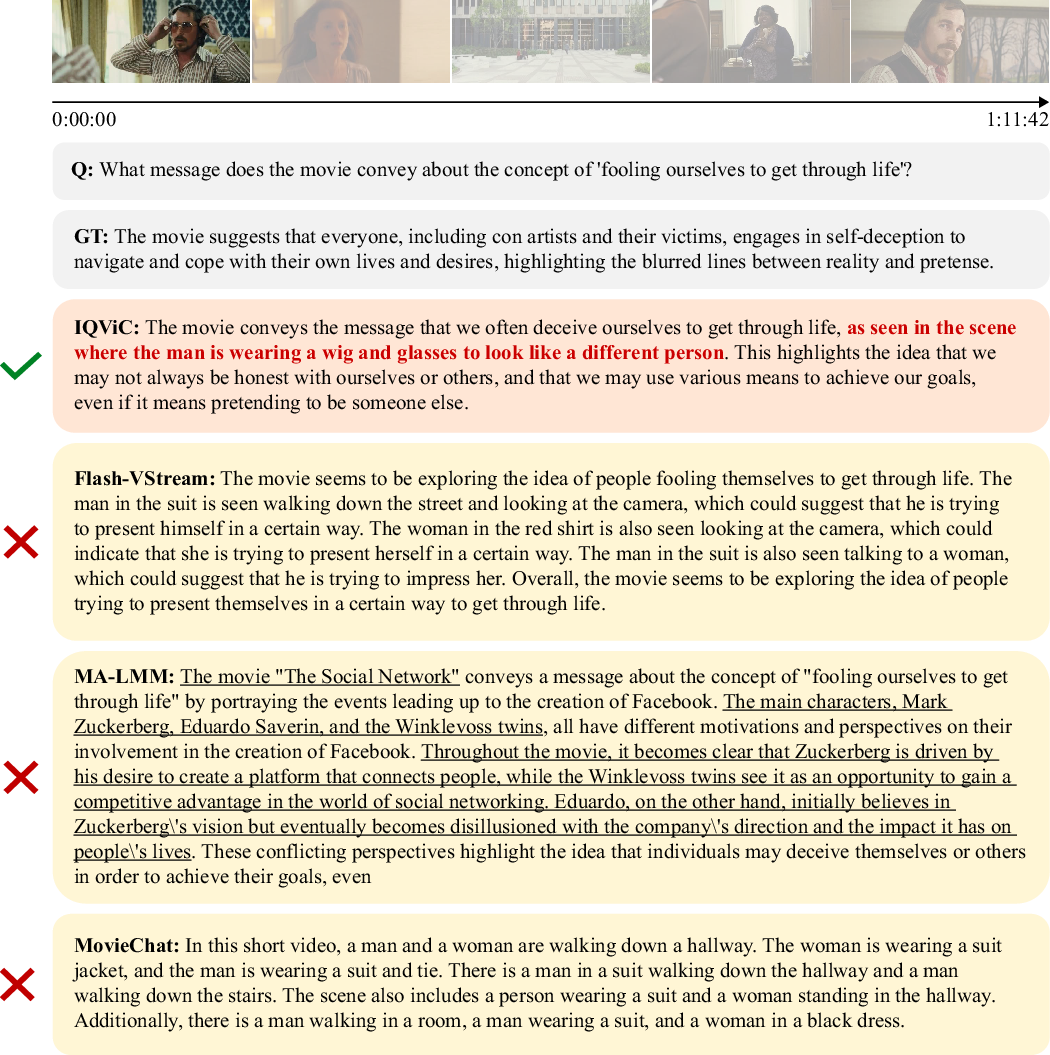}
    \caption{Question and answer examples from the InfiniBench-Vision dataset.
    The video clip is from the movie ``American Hustle" (2013), which has a duration of 1 hour and 11 minutes, 
    and the question requires understanding the deep context of the scenarios.
    The answer texts are the output exactly as generated by the methods.
    Reasonable and visually correct answers to the question are highlighted in bold.
    The answers that are clearly incorrect or hallucinated are underlined.
    }
    \label{fig:result_example}
  \end{figure*}

To better understand the performance of the proposed IQViC framework, 
we present an example of a QA pair from the InfiniBench-Vision dataset.
As shown in Figure~\ref{fig:result_example}, 
although the video is long and contains various scenes and the question requires understanding of the deep context,
the proposed method successfully answers the question because of the visual compressor and context memory mechanism,
while the other methods fail to provide the correct answer or hallucinate the answer.
MovieChat provides a partial description of the video scene but fails to answer the question.
MA-LMM clearly hallucinates and fails to provide the correct answer.
Flash-VStream attempts to reason to obtain the correct answer to the question, but the logic between the visual evidence and answer is unclear.
This provides a clear demonstration of the effectiveness of the proposed method in long-term video understanding.

\section{Conclusion}

In this paper, we propose a simple yet effective LMM framework for long-term video understanding
that incorporates a novel visual compressor, In-context, Question Adaptive Visual Compressor (IQViC).
Our approach addresses the limitations of existing memory-augmented LLM frameworks, 
which often struggle with maintaining performance over extended video sequences and handling intricate dependencies within the content.
The key innovation is the introduction of a visual compressor inspired by humans' memory and selective attention mechanisms,
which exploits transformer-based in-context compression to extract important information to answer a given question from video frames,
adapting to the given question.
Through extensive experiments for long-term and short-term video understanding, 
we demonstrate that the proposed IQViC framework outperforms state-of-the-art methods in long-term video understanding 
and achieves comparable or superior performance to conventional methods in short-term video understanding,
using a simpler memory architecture and more efficient memory token utilization than the existing methods.
Future work will focus on extending the IQViC framework to incorporate temporal information in the compression and memory mechanisms 
to enhance causal and temporal reasoning capabilities.
Another promising direction is to apply IQViC to other modalities, such as audio and 3D data.

{
    \small
    \bibliographystyle{unsrt}
    \bibliography{main}
}

\clearpage
\setcounter{page}{1}
\maketitlesupplementary

We present details of a new dataset called InfiniBench-Vision in Section~\ref{sec:infinibench-vision}. 
The limitations of the proposed method are discussed in Section~\ref{sec:limitation}. 

\section{InfiniBench-Vision}
\label{sec:infinibench-vision}

To evaluate the long-term video understanding capabilities of the proposed method,
we introduce InfiniBench-Vision, a subset of the InfiniBench~\cite{Ataallah2024InfiniBenchAC}. 
This section describes the motivation behind creating the InfiniBench-Vision dataset, its curation process, and its statistics.

\begin{figure*}[t] 
    \begin{minipage}[b]{1.0\hsize}
      \centering
      \includegraphics[width=1.0\linewidth]{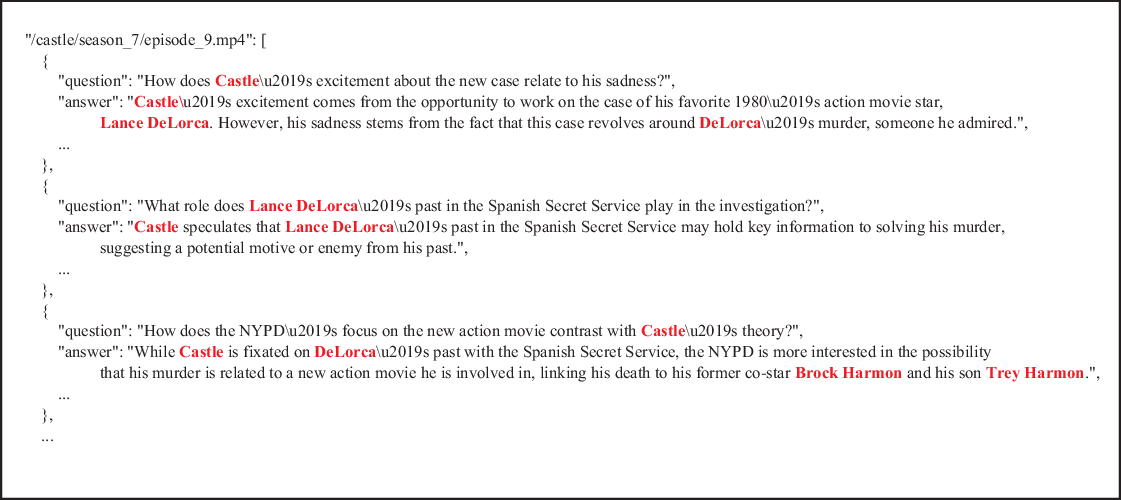}
      \subcaption{InfiniBench}
      \vspace{0.4cm}
      \label{fig:curation_before}
    \end{minipage}
    \begin{minipage}[b]{1.0\hsize}
      \centering
      \includegraphics[width=1.0\linewidth]{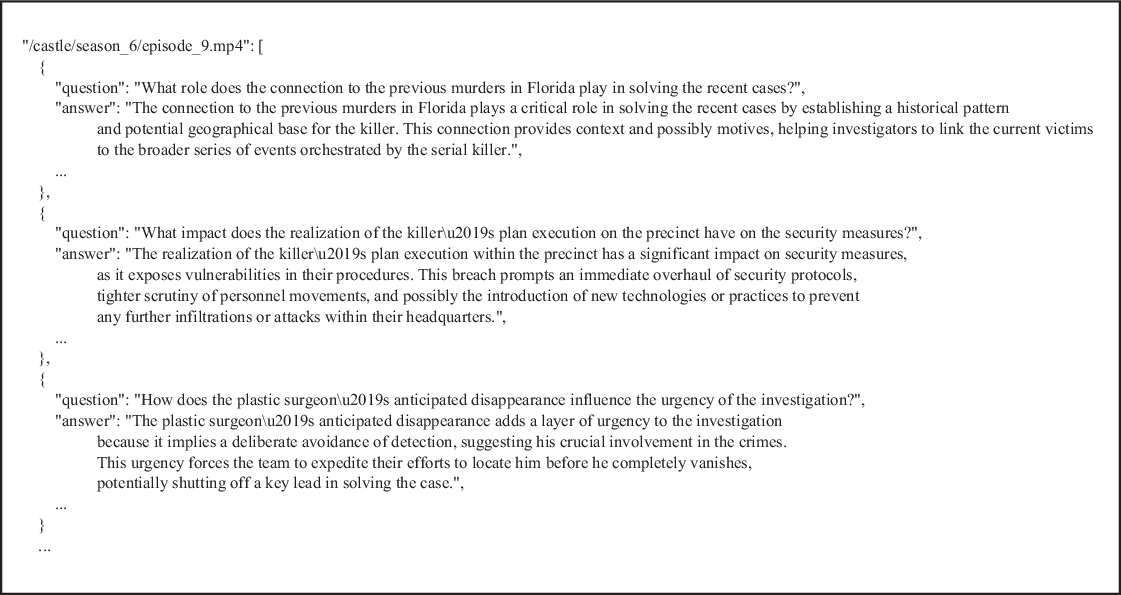}
      \subcaption{InfiniBench-Vision}
      \label{fig:curation_after}
    \end{minipage}
    \caption{Examples of QA pairs from the InfiniBench and InfiniBench-Vision annotation files. 
    Person names not identifiable solely from visual information are highlighted in bold.
    (a)~InfiniBench contains QA pairs requiring external information beyond video frames.
    (b)~InfiniBench-Vision is a subset of InfiniBench, focusing on QA pairs answerable from video frames only.
    }
    \label{fig:curation_before_after}
    \end{figure*}

\paragraph{Motivation}
Recently, several benchmarks have been proposed to evaluate the performance of long-term video 
understanding~\cite{Zhou2024MLVUAC, Fu2024VideoMMETF, Wang2024LVBenchAE, Ataallah2024InfiniBenchAC}.
However, due to commercial licensing restrictions, only InfiniBench~\cite{Ataallah2024InfiniBenchAC} was accessible for our study (Table~\ref{tab:dataset}).
InfiniBench is a large-scale, challenging long-term video understanding benchmark comprising 108k QA pairs, 
1,219 videos (averaging 53 minutes in duration), and supporting video subtitles and summaries.
While InfiniBench is a comprehensive benchmark for LMMs in long-term video understanding, 
our analysis revealed that most QA pairs require supplementary information (such as video subtitles or summaries) 
beyond video frames for accurate answers. 
Figure~\ref{fig:curation_before} illustrates that many challenging questions in InfiniBench rely on video-specific details, 
such as character names (e.g., ``Sheldon," ``Howard"), unidentifiable from video frames alone.

Our approach to long-term video understanding uses only video frames, 
similar to existing methods~\cite{Li2023LLaMAVIDAI,Maaz2023VideoChatGPTTD,Zhang2023VideoLLaMAAI, Song2023MovieChatFD,He2024MALMMML,Zhang2024FlashVStreamMR}. 
Therefore, we created a new dataset by filtering questions requiring external information from InfiniBench. 
Assuming that a significant portion of questions unanswerable from video frames alone references video-specific character names, 
we filtered these questions to create a suitable dataset, 
following the annotation policies of existing benchmarks~\cite{Zhou2024MLVUAC, Wang2024LVBenchAE}.
This resulted in InfiniBench-Vision, a subset of InfiniBench containing 265 videos (averaging 49 minutes in duration) and 599 open-ended QA pairs, 
better suited for evaluating long-term video understanding based solely on video frames (Figure~\ref{fig:curation_after}).

\begin{figure}[t]
  \centering
  \includegraphics[width=1.0\linewidth]{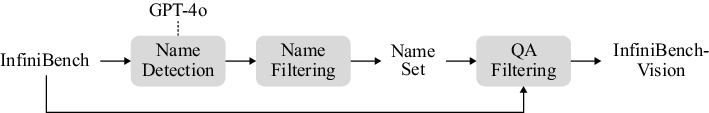}
  \caption{Curation process for the InfiniBench dataset.}
  \label{fig:curation_process}
  \end{figure}

\paragraph{Curation Process}
The curation process for the InfiniBench-Vision dataset comprises three stages, illustrated in Figure~\ref{fig:curation_process}. 

\begin{figure*}[t]
  \centering
  \includegraphics[width=1.0\linewidth]{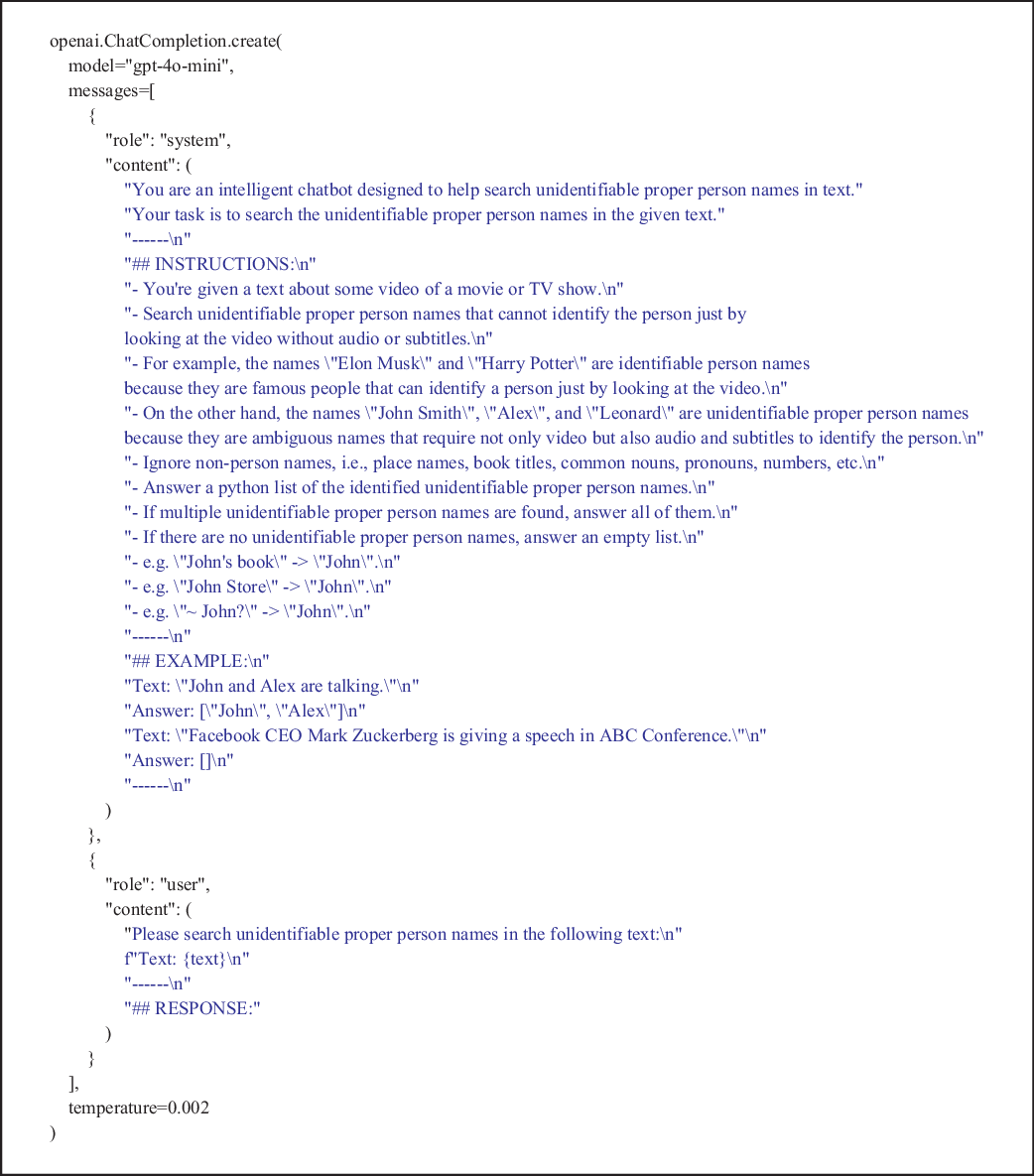}
  \caption{Prompt for GPT-4o to curate the QA pairs in the InfiniBench dataset.}
  \label{fig:prompt_curation}
  \end{figure*}

First, we identify character names within all InfiniBench QA pairs using GPT-4 (gpt-4o-mini), 
guided by a prompt specifically designed for dataset curation (Figure~\ref{fig:prompt_curation}). 
This prompt instructs GPT-4 to act as a video-specific character name detector, 
focusing on unidentifiable proper names requiring information beyond video frames (e.g., ``Sheldon," ``Howard"). 
Readily identifiable names (e.g., ``Elon Musk," ``Harry Potter") discernible from video frames alone are excluded.
This process generates a set of detected video-specific character names, 
which may include false positives such as numbers (``123") and special characters (``\#").

Second, we filter this initial set, removing the false positives based on predefined rules. 
These rules exclude non-capitalized words, articles (e.g., ``The," ``A"), numbers, and special characters. 
The resulting set contains only the video-specific character names intended for exclusion from InfiniBench. 

Finally, we extract the QA pairs that do not contain any of the refined set of character names. 
These remaining QA pairs constitute the InfiniBench-Vision dataset, 
specifically designed for evaluating long-term video understanding based solely on visual information. 

\begin{table}[t]
  \small
  \centering
  \caption{Comparison of the long-term video QA benchmarks. 
  ``Time'' indicates the average video duration in minutes. 
  ``VO'' denotes benchmarks designed for evaluation using video frames only.
  ``CU'' indicates whether commercial use is permitted.
  }
  \begin{tabular}{@{}lccccc@{}}
    \toprule
    {\bf Dataset} & {\bf \#QA} & {\bf \#Video} & {\bf Time} & {\bf VO} & {\bf CU}\\ 
    \midrule
    MLVU~\cite{Zhou2024MLVUAC} & 2.5k & 757 & 12 & \cmark & \xmark \\
    Video-MME~\cite{Fu2024VideoMMETF} & 2.7k & 900 & 17 & \xmark & \xmark \\ 
    LVBench~\cite{Wang2024LVBenchAE} & 1.5k & 103 & 68 & \cmark & \xmark \\ 
    InfiniBench~\cite{Ataallah2024InfiniBenchAC} & 108.2k & 1219 & 53 & \xmark & \cmark \\
    {\bf InfiniBench-Vision} & 0.6k & 265 & 49 & \cmark & \cmark \\ 
    \bottomrule
  \end{tabular}
  \label{tab:dataset}
  \end{table}

\paragraph{Dataset Statistics}
Table~\ref{tab:dataset} summarizes the statistics of the InfiniBench-Vision dataset. 
A rigorous curation process selected 599 QA pairs from 108k initial candidates in InfiniBench,
focusing on those suitable for evaluating long-term video understanding using video frames only (see Figure~\ref{fig:curation_before_after}). 
These QA pairs primarily are derived from the deep context understanding and multiple event linking tasks defined in InfiniBench.
The dataset comprises 265 videos of TV dramas and movies, ranging in length from 18 to 151 minutes, with an average length of 49 minutes.

\section{Limitation}
\label{sec:limitation}
This study has several limitations, which also represent exciting avenues for future research.
First, achieving accurate responses to new video questions, 
as highlighted by the ``Invisible Gorilla" experiment~\cite{Simons1999GorillasIO},
necessitates dynamic memory updates tailored to each text prompt. 
The current requirement to reprocess the entire video for each query is computationally expensive 
and impractical for online, interactive, long-video understanding applications. 
Future work will focus on mitigating this computational burden by developing novel techniques to optimize memory updates. 
This will involve assessing the relevance of incoming questions to the existing memory state 
and dynamically adjusting the update magnitude accordingly.
Second, the high VRAM consumption during training prevented end-to-end training of the video-based IQViC and the complete architecture. 
Future research will explore strategies for end-to-end training on both long and short videos to improve video comprehension.
This includes investigating techniques such as sliding windows or sampling, 
and integrating long-context vision-language models like 
LongVA's context length extension~\cite{Zhang2024LongCT} or LongVILA~\cite{Xue2024LongVILASL}.
It should be noted that this limitation applies only to the training phase; 
compression during inference yields a beneficial effect.
Finally, 
like existing methods~\cite{Li2023LLaMAVIDAI,Maaz2023VideoChatGPTTD,Zhang2023VideoLLaMAAI, Song2023MovieChatFD,He2024MALMMML,Zhang2024FlashVStreamMR},
the current method focuses solely on visual information, 
neglecting potentially valuable external knowledge sources such as subtitles, audio, and temporal information,
Integrating this information could significantly improve long-term video understanding~\cite{zou2024secondshoursreviewingmultimodal}. 
Future work will explore expanding IQViC into a multimodal information compressor 
that incorporates these external knowledge sources to enhance video comprehension.

\end{document}